\documentclass[preprint,12pt]{elsarticle}

\usepackage{amssymb}
\usepackage{amsmath}
\usepackage{float}
\usepackage{placeins} 
\usepackage{array}
\usepackage{textcomp,marvosym}
\usepackage[utf8]{inputenc}
\usepackage[T1]{fontenc}
\usepackage{float}
\usepackage{algorithm}
\usepackage{algpseudocode}
\usepackage{float}
\usepackage{booktabs}
\usepackage{placeins}
\usepackage{tabularx}
\usepackage[table,xcdraw]{xcolor}
\usepackage[justification=raggedright,singlelinecheck=false]{caption}

\usepackage{amssymb}
\usepackage{amsmath}

\journal{Computers in Biology and Medicine}

\begin{document}

\begin{frontmatter}



\title{Clinically Interpretable Mortality Prediction for ICU Patients with Diabetes and Atrial Fibrillation: A Machine Learning Approach}


\author[a]{Li Sun}
\ead{lsun4765@usc.edu}

\author[a]{Shuheng Chen}
\ead{shuhengc@usc.edu}

\author[a]{Yong Si}
\ead{yongsi@usc.edu}

\author[a]{Junyi Fan}
\ead{junyifan@usc.edu}

\author[a]{Maryam Pishgar\corref{cor1}}
\ead{pishgar@usc.edu}

\author[b]{Elham Pishgar}
\ead{dr.elhampishgar@gmail.com}

\author[c]{Kamiar Alaei}
\ead{kamiar.alaei@csulb.edu}

\author[d]{Greg Placencia}
\ead{gvplacencia@cpp.edu}

\affiliation[a]{organization={Daniel J. Epstein Department of Industrial \& Systems Engineering, University of Southern California}, 
                city={Los Angeles}, 
                state={CA}, 
                postcode={90089}, 
                country={United States}}

\affiliation[b]{organization={Colorectal Research Center, Iran University of Medical Sciences}, 
                city={Tehran}, 
                postcode={14535}, 
                country={Iran}}

\affiliation[c]{organization={Department of Health Science, California State University}, 
                city={Long Beach}, 
                state={CA}, 
                postcode={90840}, 
                country={United States}}

\affiliation[d]{organization={Department of Industrial and Manufacturing Engineering, California State Polytechnic University}, 
                city={Pomona}, 
                state={CA}, 
                postcode={91768}, 
                country={United States}}

\cortext[cor1]{Corresponding author. Email: \texttt{pishgar@usc.edu}}

\begin{abstract}
\textbf{Background:} Patients with both diabetes mellitus (DM) and atrial fibrillation (AF) face elevated mortality in intensive care units (ICUs), yet models targeting this high-risk group remain limited.

\textbf{Objective:} To develop an interpretable machine learning (ML) model predicting 28-day mortality in ICU patients with concurrent DM and AF using early-phase clinical data.

\textbf{Methods:} A retrospective cohort of 1,535 adult ICU patients with DM and AF was extracted from the MIMIC-IV database. Data preprocessing involved median/mode imputation, z-score normalization, and early temporal feature engineering. A two-step feature selection pipeline—univariate filtering (ANOVA F-test) and Random Forest-based multivariate ranking—yielded 19 interpretable features. Seven ML models were trained with stratified 5-fold cross-validation and SMOTE oversampling. Interpretability was assessed via ablation and Accumulated Local Effects (ALE) analysis.

\textbf{Results:} Logistic regression achieved the best performance (AUROC: 0.825; 95\% CI: 0.779–0.867), surpassing more complex models. Key predictors included RAS, age, bilirubin, and extubation. ALE plots showed intuitive, non-linear effects such as age-related risk acceleration and bilirubin thresholds.

\textbf{Conclusion:} This interpretable ML model offers accurate risk prediction and clinical insights for early ICU triage in patients with DM and AF.
\end{abstract}

\begin{graphicalabstract}
\includegraphics[width=\textwidth]{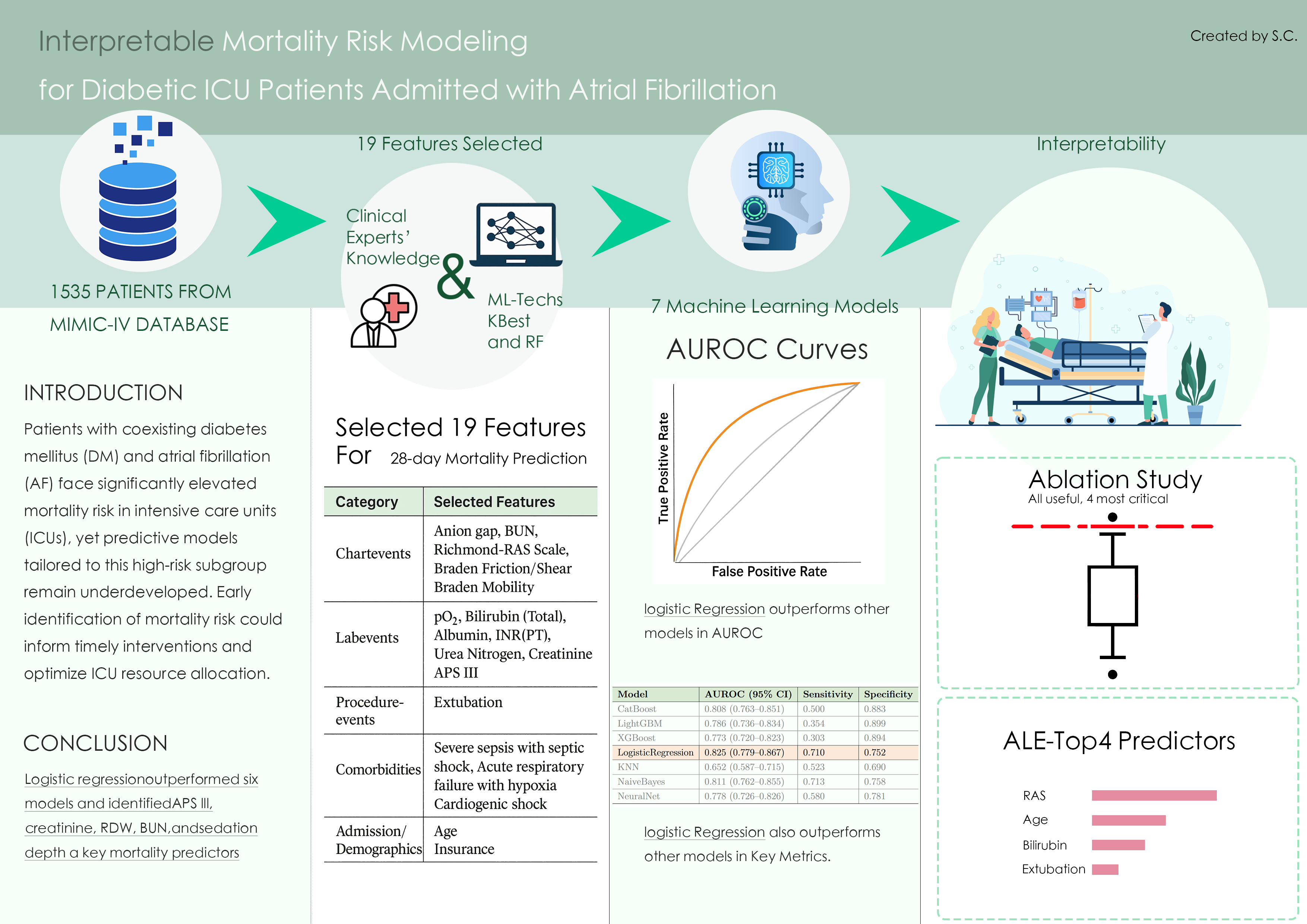}
\end{graphicalabstract}

\begin{highlights}
\item A two-stage feature selection pipeline combining ANOVA F-test and Random Forest importance yielded 19 clinically relevant predictors across five domains.
\item Class imbalance was addressed using SMOTE within a stratified 5-fold cross-validation framework, enhancing model stability and generalization.
\item Logistic regression outperformed six alternative models, achieving the highest test AUROC (0.825) and offering superior clinical interpretability.
\item ALE, and ablation analysis identified  Richmond-RAS Scale, age, total bilirubin, and extubation status as key mortality predictors, enabling bedside decision support.
\end{highlights}


\end{frontmatter}



\section{Introduction}
\label{sec1}
Diabetes mellitus (DM) is a chronic metabolic disorder characterized by persistent hyperglycemia due to defects in insulin secretion, insulin action, or both, and has become a major public health issue worldwide\cite{eisenbarth1986type,skyler2004diabetes,american2010diagnosis,kaveeshwar2014current}. According to the International Diabetes Federation (IDF), as of 2024, approximately 589 million adults aged 20–79 years are living with diabetes globally, representing 1 in 9 individuals within this age group. Projections indicate that this number will escalate to 853 million by 2050, marking a 45\% increase\cite{idf2024}. Concurrently, global health expenditure related to diabetes has surpassed USD 1 trillion for the first time, reflecting a 338\% increase over the past 17 years . These trends underscore the escalating public health and economic challenges posed by diabetes worldwide\cite{whiting2011idf,saeedi2019global}.

Long recognized for its profound metabolic disturbances, diabetes mellitus is now equally noted for its extensive complications, which substantially elevate both morbidity and mortality rates worldwide\cite{tomic2022burden,young2008diabetes,raghavan2019diabetes}. Among these complications, atrial fibrillation (AF) stands out as a frequent and particularly serious comorbidity in diabetic populations. Epidemiological studies have demonstrated that individuals with diabetes are at a significantly higher risk of developing AF compared to those without diabetes\cite{seyed2020risk,aune2018diabetes,alwafi2020epidemiology,wang2019atrial}. More importantly, the coexistence of AF and diabetes confers a markedly increased risk of adverse cardiovascular outcomes—including stroke, systemic embolism, heart failure, and, most critically, all-cause and cardiovascular mortality—compared to having either condition alone\cite{wang2019atrial,hu2021impact,xu2022impact}. The amplified mortality risk observed in individuals with coexisting atrial fibrillation and diabetes mellitus likely results from interconnected pathological mechanisms, including enhanced systemic inflammation, increased thrombotic potential, and accelerated cardiovascular and renal damage\cite{guo2012inflammation,ding2020atrial,morel2017incidence}. Given the rising global prevalence of both diabetes and AF, and the grave prognosis associated with their coexistence, there is a pressing need for early and accurate prediction of mortality risk in this vulnerable population, especially those in Intensive Care Units (ICUs). Effective risk prediction not only helps identify high-risk patients, but also enables clinicians to implement tailored interventions and ultimately improve survival outcomes. Therefore, developing robust predictive models for mortality in ICU patients with both AF and diabetes is a critical step toward optimizing clinical management and reducing the global burden of these conditions.

Despite the clinical urgency, few studies to date have specifically addressed mortality prediction in critically ill patients with coexisting atrial fibrillation and diabetes mellitus. Most existing research either focuses on AF or diabetes separately, or includes these conditions as covariates rather than primary targets in broader ICU prognostic models\cite{seyed2020risk,aune2018diabetes,karayiannides2018high,huang2022diabetes}. This gap may be attributed to the traditionally fragmented approach of studying chronic comorbidities in isolation and the relatively limited representation of AF-DM co-diagnosed patients in public ICU databases. For example, Karayiannides et al. (2018) conducted a comprehensive nationwide cohort study in Sweden, encompassing over 320,000 patients with non-valvular AF\cite{karayiannides2018high}. Their findings demonstrated that individuals with both AF and diabetes mellitus exhibited significantly higher rates of adverse outcomes, including mortality, heart failure, ischaemic stroke, myocardial infarction, and bleeding, compared to those without diabetes. Another example is a recent analysis from the prospective GLORIA-AF registry by Liu et al. (2024), they evaluated over 15,000 individuals with AF to explore the impact of DM and insulin use on adverse clinical outcomes\cite{liu2024gloria}. Their findings revealed that patients with both AF and DM experienced significantly increased risks of all-cause mortality, cardiovascular death, major bleeding, myocardial infarction (MI), and major adverse cardiovascular events (MACE) compared to those without DM. Similarly, Geng et al. (2022) investigated the longitudinal impact of new-onset AF on adverse health outcomes in a cohort of over 16,000 adults with type 2 diabetes (T2D) from the UK Biobank\cite{geng2022onsetaf}. Excluding individuals with pre-existing cardiovascular disease (CVD) and chronic kidney disease (CKD) at baseline, their study demonstrated that patients with T2D who subsequently developed AF were at significantly elevated risk for atherosclerotic cardiovascular disease (ASCVD), heart failure, CKD, all-cause mortality, and cardiovascular mortality compared to those who remained free of AF. Over a median follow-up of approximately 11 years, incident AF was associated with a nearly twofold higher risk of ASCVD, more than fourfold risk of heart failure, and significantly increased risks of CKD and death. These findings emphasize the critical role of AF prevention in mitigating both macrovascular and microvascular complications among patients with T2D. 

Although these studies underscore the significant prognostic implications of this comorbidity, they fall short in providing intensive care clinicians with practical, early-stage assessment tools necessary for predicting patient mortality risk using initial clinical information. In this context, machine learning (ML) approaches offer a promising solution, as they can efficiently integrate complex, high-dimensional data from the early phase of ICU admission and generate accurate, individualized risk predictions in real time.

In recent years, ML methods have gained prominence in clinical outcome prediction, particularly within ICU settings. For example, Chen et al. (2025) developed an XGBoost-based model to predict ICU readmission in acute pancreatitis patients using the MIMIC-III database. The model demonstrated strong discriminative ability, achieving an Area Under the Receiver Operating Characteristic curve (AUROC) of 0.862 (95\% Confidence Interval (CI): 0.800–0.920), and identified platelet count, age, and SpO2 as key predictors\cite{chen2025predicting}. Fan et al. (2025) developed a LightGBM-based model to predict 28-day mortality in ICU patients with heart failure and hypertension using the MIMIC-IV database. The model achieved an AUROC of 0.8921 (95\% CI: 0.8694–0.9118) and retained strong performance in external validation (AUROC: 0.7404), highlighting its potential to improve early risk stratification and guide personalized interventions in this high-risk population\cite{fan2025predicting}. Ashrafi et al. (2024) developed a Random Forest-based model to predict sepsis outcomes in ICU patients using the MIMIC-IV database. The model achieved an AUROC of 0.94 (±0.01) and identified SOFA score and urine output as key predictors through SHapley Additive exPlanations (SHAP) analysis, offering a high-performing and interpretable tool to support clinical decision-making in sepsis management\cite{gao2024prediction}. Li et al. (2024) developed a logistic regression (LR)-based model to predict postoperative pneumonia in patients with aneurysmal subarachnoid hemorrhage using perioperative data. The model achieved an AUC of 0.91 in internal validation and 0.89 in external validation, with key predictors including mechanical ventilation time, GCS, and inflammatory markers, supporting early risk identification and targeted prevention\cite{li2024development}. By accommodating complex interactions among high-dimensional variables, ML models can outperform traditional statistical approaches and facilitate individualized risk assessment. Moreover, the integration of interpretable ML techniques—such as Accumulated Local Effects (ALE)—enables clinicians to identify key predictive features and gain actionable insights from model outputs \cite{lundberg2017unified}. These advantages render ML particularly suitable for understanding and predicting outcomes in heterogeneous, high-risk populations such as AF-DM patients. Among these ML models, LR remains a widely adopted baseline model due to several key advantages: (1) Interpretability. LR provides direct estimates of variable effects, facilitating clinical understanding and transparency in decision-making; (2) Stability with limited data. LR performs well in small to moderate-sized datasets, a common scenario in specific clinical subgroups; and (3) Robustness in binary classification. LR is well-suited for predicting binary outcomes and, when regularized, demonstrates strong generalizability in high-dimensional clinical settings. These strengths make LR a reliable tool for early risk stratification in critical care populations. 

This study introduces methodological and clinical innovations that distinguish our work from prior ICU mortality prediction research and directly address urgent unmet needs for patients with coexisting diabetes and atrial fibrillation:
\begin{itemize}
\item To our knowledge, this is the first study to develop an interpretable, early-phase machine learning model for 28-day mortality specifically tailored to adult ICU patients with both diabetes mellitus and atrial fibrillation, leveraging real-world data from the MIMIC-IV database. 
\item We designed and implemented a two-stage hybrid feature selection pipeline, combining univariate ANOVA F-test with multivariate Random Forest Gini importance, to refine over 500 candidate variables into a set of 19 clinically actionable predictors. This strategy ensures both statistical robustness and high clinical interpretability, spanning demographic, laboratory, procedural, severity, and comorbidity domains.
\item To address the marked class imbalance in ICU mortality, we uniquely incorporated SMOTE-based oversampling strictly within a stratified 5-fold cross-validation framework, enhancing generalizability and model stability while minimizing information leakage—a step not commonly performed in prior ICU ML studies.
\item Through systematic head-to-head comparison of seven machine learning algorithms, we demonstrated that regularized logistic regression—not only outperformed complex ensemble models on test AUROC (0.825, 95\% CI: 0.779–0.867) but also provided the optimal balance of predictive accuracy, calibration, and bedside interpretability, facilitating immediate clinical translation.
\item We advanced model transparency and bedside applicability via integrated ablation analysis and ALE interpretation, which collectively identified  richmond-ras scale, age, total bilirubin, and extubation status as dominant mortality predictors. These interpretable insights enable actionable, pathophysiologically coherent risk stratification at the individual patient level.
\end{itemize}

\section{Methodology}
\subsection*{Data Source and study design}

This study was conducted using the Medical Information Mart for Intensive Care IV (MIMIC-IV, version 2.2) database~\cite{johnson2020mimic}, a publicly available and de-identified electronic health record repository developed by the Massachusetts Institute of Technology in collaboration with the Beth Israel Deaconess Medical Center. MIMIC-IV contains over 60,000 ICU admissions between 2008 and 2019, spanning various care units and clinical domains including vital signs, laboratory results, procedures, medications, diagnoses, and outcomes.

The database offers high temporal resolution and clinical granularity, enabling the development of complex models grounded in real-world ICU practice. Its open-access nature also ensures transparency and reproducibility in machine learning research. All analyses in this study complied with the data use agreement for MIMIC-IV, and no further IRB approval was required due to the use of de-identified data.

To predict 28-day mortality in ICU patients with coexisting diabetes mellitus and AF, we designed a structured and interpretable machine learning pipeline that integrates clinical domain knowledge, robust statistical modeling, and state-of-the-art explainability techniques. The pipeline includes seven major stages: cohort construction, data preprocessing, feature selection, class balancing, model development, evaluation, and interpretability analysis.

From over 500 candidate variables spanning five domains (chartevents, labevents, procedureevents, comorbidities, demographics), variables were filtered based on completeness, variance, and clinical relevance. Imputation and normalization techniques were applied to prepare the data for machine learning. Features were selected through a two-step process combining univariate statistical filtering and multivariate model-based ranking. To address the substantial class imbalance in mortality outcomes, SMOTE was applied within training folds of cross-validation.

Seven machine learning models—including gradient boosting machines, linear and probabilistic classifiers, and shallow neural networks—were trained and evaluated using stratified 5-fold cross-validation and independent holdout testing. Model performance was assessed using AUROC and calibrated through 2,000 bootstrap replicates. Global and local interpretability was achieved through t-test ablation, and ALE analysis.

The complete pipeline is summarized in Algorithm~\ref{alg:mortality_prediction}, ensuring transparency, reproducibility, and clinical applicability.

\begin{algorithm}[H]
\caption{ML Pipeline for 28-Day Mortality Prediction in ICU Patients with Diabetes and Atrial Fibrillation}
\label{alg:mortality_prediction}
\begin{algorithmic}[1]
\Require MIMIC-IV data with confirmed diabetes and AF diagnosis
\Ensure Binary prediction: 28-day mortality (0 = survived, 1 = died)

\State \textbf{Step 1: Cohort Selection}
\State Identify patients using ICD-9/10 codes for diabetes and AF
\State Exclude patients $<$ 18 years, with malignancy, or multiple ICU stays
\State Retain first ICU admission only

\State \textbf{Step 2: Data Preprocessing}
\ForAll{features in dataset}
    \If{numerical}
        \State compute mean, CV, IQR; impute with median
    \ElsIf{categorical}
        \State mode imputation; apply label or one-hot encoding
    \EndIf
\EndFor
\State Drop variables with $>$80\% missingness
\State Normalize continuous variables using z-score

\State \textbf{Step 3: Feature Selection}
\State Apply ANOVA F-test to select top features (SelectKBest)
\State Rank using Random Forest Gini importance
\State Retain top 19 interpretable features

\State \textbf{Step 4: Class Imbalance Handling}
\State Apply SMOTE oversampling in training folds only

\State \textbf{Step 5: Model Training and Tuning}
\ForAll{model $\in$ \{CatBoost, XGBoost, LightGBM, Logistic Regression, KNN, Naive Bayes, NeuralNet\}}
    \State Grid search over model-specific hyperparameters
    \State Evaluate on training folds using AUROC and F1
\EndFor
\State Evaluate best model on test set (holdout 30\%)

\State \textbf{Step 6: Statistical and Interpretability Analysis}
\State Use t-tests to compare training and test distributions
\State Perform ablation to assess feature contribution
\State Apply ALE to interpret localized, non-linear effects

\end{algorithmic}
\end{algorithm}

\subsection*{Patient Selection}

This study utilized the MIMIC-IV database to identify a clinically meaningful cohort of ICU patients with a documented history of diabetes mellitus who were hospitalized due to AF, aiming to develop an early mortality risk prediction model. A structured cohort selection process was implemented to ensure clinical homogeneity and to align with the goal of early risk stratification in critically ill diabetic patients with AF.

We first restricted the dataset to include only each patient's first ICU admission (n = 35,794) to avoid duplication and ensure that the clinical variables reflected the initial state of critical illness. Within this subset, patients with a confirmed diagnosis of diabetes mellitus were identified using standardized ICD-9/10 codes (n = 6,251), given the known impact of diabetes on ICU prognosis.

To capture the specific clinical scenario of interest, we further filtered the cohort to include only those patients who also had a diagnosis of atrial fibrillation (n = 2,258). Patients under 18 years of age were excluded (n = 1,834) to eliminate the confounding effect of pediatric care protocols. Additionally, individuals with active malignancy or metastatic cancer were removed (n = 1,535), as their disease trajectory and treatment goals differ significantly from the general ICU population.

The final study cohort comprised 1,535 adult ICU patients with coexisting diabetes and atrial fibrillation, free from malignancy and undergoing their first ICU admission. For each individual, structured clinical variables—including demographics, comorbidities, vital signs, laboratory values, and interventions—were collected from the first 48 hours of ICU stay to enable early-phase mortality prediction. The primary outcome of interest was in-hospital mortality, defined based on hospital discharge status.

\begin{figure}[H]
\centering
\includegraphics[width=1\linewidth]{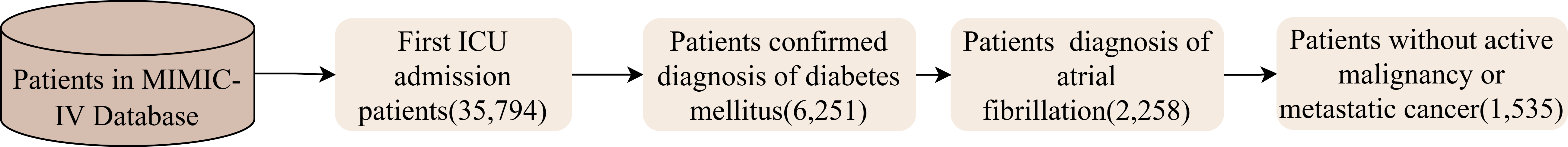} 
\caption{\textbf{Flowchart Illustrating the Patient Selection Process from MIMIC-IV}}
\label{fig:patient_selection}
\end{figure}

A graphical summary of the patient selection process is shown in Figure \ref{fig:patient_selection}.

\subsection*{Data Preprocessing}

To ensure data integrity and optimize the analytical value of the cohort, we implemented a comprehensive preprocessing pipeline tailored to diabetic patients admitted to the ICU with AF. This multi-step strategy addressed the inherent challenges of ICU data—such as missingness, variable scales, temporal instability, and class imbalance—while preserving the clinical interpretability essential for mortality risk prediction.

Missing data were treated separately for numerical and categorical variables~\cite{shi2025prediction}. Numerical features such as \textit{Urea Nitrogen}, \textit{pO\textsubscript{2}}, and \textit{Creatinine}, which are relevant to renal and respiratory dysfunction, were imputed using median values to reduce the influence of skewed distributions. Categorical features including \textit{Braden Mobility}, \textit{Extubation status}, and \textit{Richmond-RAS Scale} were imputed using mode values to reflect the most prevalent clinical states~\cite{Chen2025}.

To quantify within-day physiological fluctuations, we calculated the coefficient of variation (CV) and interquartile range (IQR) for time-series variables such as \textit{Anion gap}, \textit{Peak Inspiratory Pressure}, and \textit{apsiii} during the first 48 hours of ICU admission:

\begin{equation}
\text{CV}(x_i) = \frac{\sigma(x_i)}{\mu(x_i)}, \quad 
\text{IQR}(x_i) = Q_3(x_i) - Q_1(x_i)
\label{eq:temporal_stats_af}
\end{equation}

Here, $\mu(x_i)$ and $\sigma(x_i)$ represent the mean and standard deviation of feature $x_i$, while $Q_1$ and $Q_3$ denote its 25th and 75th percentiles. These metrics enabled the model to capture acute physiologic variability, often indicative of decompensation in AF patients.

All continuous variables were standardized using z-score normalization:

\begin{equation}
z_i = \frac{x_i - \mu}{\sigma}
\label{eq:z_score_af}
\end{equation}

This transformation ensured numerical stability and comparability across variables for scale-sensitive models such as gradient boosting classifiers.

To address the significant class imbalance in 28-day mortality outcomes, SMOTE was employed within the training folds of stratified 5-fold cross-validation~\cite{hassanzadeh2023hospital}. This technique synthetically generated new samples for the minority class (those who died within 28 days of ICU admission) by interpolating between neighboring samples. Importantly, all transformations—including imputation, scaling, encoding, and resampling—were applied only to the training set to prevent data leakage, with validation and test sets left untouched.

In summary, this preprocessing strategy facilitated robust, interpretable modeling of 28-day mortality risk following ICU admission in diabetic patients with AF.

\subsection*{Feature Selection}

We began with an initial pool of over 500 candidate variables from the MIMIC-IV database, each screened based on data availability, prevalence, and clinical documentation standards. This initial list was refined through structured filtering: variables with excessive missingness ($>$20\%) or minimal occurrence (documented in fewer than 100 patients) were excluded prior to formal analysis.

The feature preselection process was conducted under the guidance of experienced critical care specialists, whose domain expertise ensured that the retained variables were clinically plausible and relevant to real-world ICU practice.

To facilitate clinically grounded feature engineering and enhance interpretability in downstream modeling, we organized all candidate variables into five structured categories based on their data provenance and physiologic relevance: chartevents, labevents, procedureevents, comorbidities, and administrative/demographic data. This categorization was informed by the known clinical trajectories of diabetic patients with AF in the ICU, whose mortality risk is jointly shaped by hemodynamic instability, metabolic stress, organ failure, and care disparities\cite{zuo2025joint}.

The chartevents domain comprised real-time bedside assessments frequently documented by nursing and respiratory staff. Key variables such as Braden Mobility, Richmond-RAS Scale, Anion Gap, and Peak Inspiratory Pressure were retained due to their relevance in capturing immobility risk, sedation depth, metabolic acidosis, and ventilatory mechanics—critical dimensions in AF patients who are often at heightened risk of respiratory compromise and neurologic fluctuation\cite{staudacher2024impact}.

The labevents category included core laboratory parameters that provided insights into systemic dysfunction across renal, hepatic, and coagulation axes. Variables such as Urea Nitrogen, Creatinine, and Albumin served as surrogates for renal filtration and nutritional status\cite{richter2019blood}, while INR(PT) and Total Bilirubin reflected hepatic synthetic function and cholestasis. Additionally, the inclusion of APS III—a composite illness severity index—enabled quantification of global physiologic burden within the first 48 hours of ICU stay\cite{lin2024global}.

Procedureevents were used to capture the occurrence and timing of invasive interventions, with Extubation emerging as a critical temporal marker of respiratory trajectory\cite{backman2023lung}. In the context of AF, where fluctuations in hemodynamics and consciousness are common, extubation status reflects both weaning potential and treatment intensity\cite{baptistella2018predictive}.

The comorbidities domain included ICD-based diagnostic flags denoting critical illness complexity. Conditions such as Severe sepsis with septic shock, Acute respiratory failure with hypoxia, and Cardiogenic shock were retained based on their mechanistic relevance and established association with increased short-term mortality in AF and diabetic populations\cite{storgaard2013short}. These diagnoses reflect systemic insults that may synergize with arrhythmic burden and metabolic dysregulation.

Lastly, administrative and demographic features—notably Age and Insurance Type—were preserved to account for baseline vulnerability and structural disparities in care access. These variables, while non-biological, are increasingly recognized as key modifiers of ICU outcomes, especially in chronically ill populations.

To refine this broad and heterogeneous candidate pool, we employed a two-stage hybrid feature selection strategy that combined univariate statistical filtering with multivariate model-based refinement—thereby balancing computational efficiency, statistical robustness, and clinical interpretability.

In the first stage, we applied the ANOVA F-test using the SelectKBest method to evaluate the marginal relevance of each feature to the binary outcome of 28-day mortality following ICU admission\cite{zollanvari2023feature}:

\begin{equation}
F(x_i) = \frac{\text{Between-group variance of } x_i}{\text{Within-group variance of } x_i}
\label{eq:anova_af}
\end{equation}

This method, well-suited for high-dimensional clinical data, identifies features that exhibit significant group-level separation between survivors and non-survivors. By assessing each feature independently, it provides an efficient means of eliminating uninformative or weakly correlated variables prior to model training. At this stage, we removed the following low-ranking variables due to minimal discriminatory power: \textit{Temperature}, \textit{Glasgow Coma Scale (GCS)}, \textit{SpO\textsubscript{2}}, . These features either lacked statistical association with the outcome or exhibited significant overlap across survival groups, limiting their utility in early-phase discrimination.

In the second stage, we employed a Random Forest classifier to evaluate multivariate feature importance using Gini impurity reduction\cite{parmar2018review}:

\begin{equation}
I(x_i) = \sum_{t \in T} \frac{p(t) \cdot \Delta i(t)}{f(t)}
\label{eq:gini_af}
\end{equation}

Here, \( I(x_i) \) denotes the cumulative contribution of feature \( x_i \) across all tree splits \( t \), where \( p(t) \) is the proportion of samples reaching node \( t \), \( \Delta i(t) \) is the reduction in impurity, and \( f(t) \) is the frequency of feature usage. This model-based approach captures nonlinear interactions and high-order dependencies common in ICU datasets, allowing us to further refine the feature set by identifying redundancy and contextual irrelevance.

During this stage, additional features were excluded due to low multivariate contribution or collinearity with stronger predictors. These included \textit{Lactate}, which was highly correlated with \textit{Anion gap}; \textit{ED duration}, which introduced administrative noise; and \textit{Chloride}, which offered limited incremental information beyond renal function markers. This pruning helped to minimize overfitting while preserving the diversity and clinical richness of the feature space.

Ultimately, this dual-pronged approach enabled the retention of variables that were both statistically robust and clinically interpretable. The final feature set represented a balanced cross-section of physiologic domains—including renal and hepatic function, sedation level, respiratory support, illness severity, and comorbidity burden—each contributing uniquely to 28-day mortality risk in diabetic ICU patients with atrial fibrillation.

The final selection resulted in 19 features across five domains. These included physiologic instability markers (e.g., \textit{Anion gap}, \textit{Peak Inspiratory Pressure}), severity scores (e.g., \textit{APS III}), biochemical indicators (e.g., \textit{Creatinine}, \textit{Albumin}), ICU procedures (e.g., \textit{Extubation}), and critical comorbidities (e.g., \textit{Cardiogenic shock}).

Importantly, the final list of predictors was reviewed and refined in collaboration with senior clinicians with expertise in both critical care and translational research. Their input helped ensure that the selected features not only met statistical criteria but also captured practical diagnostic, therapeutic, and prognostic considerations relevant to ICU decision-making.

\begin{table}[H]
\centering
\caption{\textbf{Final 19 Features Used for 28-Day Mortality Prediction in Diabetic ICU Patients with Atrial Fibrillation}}
\label{tab:final_features_af}
\small
\renewcommand{\arraystretch}{1.2}
\begin{tabularx}{\textwidth}{>{\raggedright\arraybackslash}p{4.5cm}|>{\raggedright\arraybackslash}X}
\hline
\rowcolor[HTML]{D9EAD3}
\textbf{Category} & \textbf{Selected Features} \\
\hline
Chartevents & Anion gap, BUN, Richmond-RAS Scale, Braden Friction/Shear, Braden Mobility, Peak Inspiratory Pressure \\
\hline
Labevents & pO\textsubscript{2}, Bilirubin (Total), Albumin, INR(PT), Urea Nitrogen, Creatinine, APS III \\
\hline
Procedureevents & Extubation \\
\hline
Comorbidities & Severe sepsis with septic shock, Acute respiratory failure with hypoxia, Cardiogenic shock \\
\hline
Admission/Demographics & Age, Insurance \\
\hline
\end{tabularx}
\end{table}

This curated, interpretable feature set formed the foundation for downstream modeling, enabling robust prediction of 28-day mortality in diabetic ICU patients with AF.

\subsection*{Model Development}

\begin{figure}[htbp]
    \centering
    \includegraphics[width=1\linewidth]{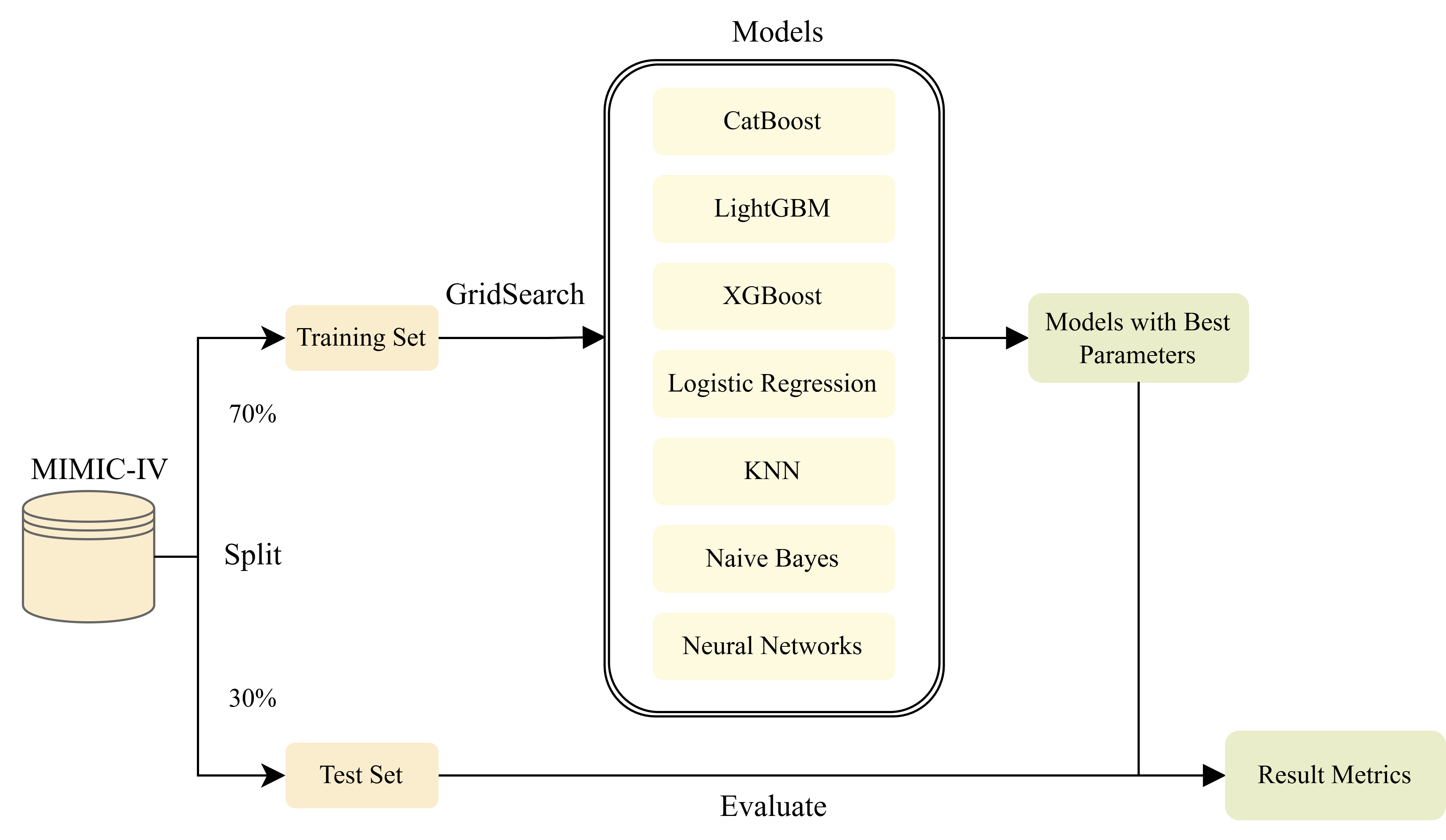}  
    \caption{\textbf{Workflow of Multi-model Development and Evaluation.}}
    \label{fig:model}
\end{figure}

To predict 28-day mortality in diabetic ICU patients with atrial fibrillation, we constructed a supervised machine learning pipeline incorporating diverse algorithmic paradigms. Following feature engineering and cohort partitioning (70\% training, 30\% testing via stratified random sampling), seven representative classifiers were implemented and optimized via stratified 5-fold cross-validation with grid search (Figure~\ref{fig:model}).

\textbf{Gradient boosting models} (CatBoost, LightGBM, XGBoost) were employed to capture non-linear relationships and interaction effects across high-dimensional, partially sparse clinical data. These ensemble learners iteratively combine weak decision trees into strong classifiers, making them well-suited for heterogeneous ICU datasets. CatBoost was selected for its ability to handle categorical features natively and mitigate overfitting via ordered boosting and symmetric tree construction. LightGBM and XGBoost, both efficient gradient boosting frameworks, utilized leaf-wise and level-wise growth strategies, respectively, with hyperparameters tuned for learning rate, tree depth, number of estimators, subsample fraction, and regularization penalties (\texttt{reg\_alpha}, \texttt{reg\_lambda}). These models provided strong discriminative performance and robust generalization under class imbalance.

\textbf{Linear models} were represented by Logistic Regression, which served as a benchmark for interpretability and transparency. L1 (Lasso) and L2 (Ridge) regularizations were used to address multicollinearity and reduce model complexity, with the inverse regularization strength \( C \) as the key tuning parameter. Despite its simplicity, logistic regression offered valuable clinical utility by providing directly interpretable feature coefficients.

\textbf{Distance-based and probabilistic models}, including k-Nearest Neighbors (KNN) and Naïve Bayes, were implemented to evaluate performance under strong structural assumptions. KNN modeled local similarity in feature space using Euclidean and Manhattan distances, with neighbor count \( k \) and weighting scheme tuned. Naïve Bayes assumed conditional independence among features and estimated class-conditional likelihoods under a Gaussian distribution. These models required minimal training time and served as efficient baselines; however, their assumptions limited flexibility in capturing complex ICU interactions.

\textbf{Neural networks} were introduced to assess the utility of flexible function approximators in modeling non-linear and high-order dependencies. A shallow feedforward architecture with a single hidden layer was implemented using the ReLU activation and sigmoid output. Key parameters included the number of hidden units, learning rate, batch size, and dropout rate. The network was optimized using the Adam optimizer and binary cross-entropy loss. While neural networks offered the highest modeling flexibility, their susceptibility to overfitting and lower interpretability posed trade-offs in a clinical setting.

All models were trained with the goal of maximizing classification accuracy while minimizing overfitting. Regularization techniques—including L1/L2 penalties, early stopping, dropout, and tree constraints—were applied where applicable. Model performance was evaluated using the area under AUROC, a threshold-independent metric quantifying discriminative ability. To quantify statistical uncertainty, 2,000 bootstrap replicates were used to compute 95\% confidence intervals on test AUROC.

This ensemble of models collectively spanned a range of complexity, interpretability, and computational cost. Their comparative performance offered a comprehensive view of algorithmic suitability for mortality risk stratification in this clinically complex ICU cohort.

\subsection*{Statistical Analyses}

To ensure the robustness, clinical interpretability, and trustworthiness of our predictive modeling framework, we implemented a multi-tiered suite of statistical analyses and model explanation techniques. Each method was selected to address a different aspect of model validity—from dataset integrity and feature contribution to individualized risk interpretation.

We began by evaluating the comparability between the training and test cohorts through descriptive statistics and inferential hypothesis testing. Two-sided t-tests (with Welch’s correction when appropriate) were applied to key continuous variables such as age, and laboratory values to assess distributional equivalence. This step verified that the stratified data split preserved the underlying population structure, thereby mitigating sampling bias and improving the generalizability of predictive performance across diverse ICU subgroups.

To assess the marginal utility of individual features, we performed an ablation analysis by iteratively excluding one feature at a time and retraining the model, quantifying the resulting change in AUROC. This procedure enabled us to isolate each variable's unique contribution to the predictive signal and provided an interpretable, performance-oriented validation of our final feature set.

To move beyond additive global explanations and capture nuanced, non-linear, and localized interactions between features and outcomes, we incorporated ALE as the core interpretability technique in this study. Unlike traditional partial dependence plots (PDPs), which can suffer from feature correlation bias and extrapolation artifacts, ALE accounts for conditional feature distributions and computes the average model response over localized intervals\cite{souli2000ale}. Mathematically, the ALE function for a feature \( x_j \) is defined as:

\begin{equation}
ALE_j(z) = \int_{z_0}^{z} \mathbb{E}_{X_{\setminus j}} \left[ \frac{\partial f(X)}{\partial x_j} \Big| x_j = s \right] ds,
\label{eq:ale}
\end{equation}

where \( f(X) \) is the model prediction function, \( x_j \) is the feature of interest, and the integration is taken over the feature’s distribution. ALE provides a reliable, model-agnostic quantification of how changes in a single predictor impact predicted risk, while properly accounting for the joint distribution of other features.

In this study, ALE plots were employed as the primary interpretability technique to elucidate clinically salient and potentially non-monotonic associations between model predictors and mortality risk. Unlike SHAP values, which provide additive, instance-level attributions ideal for assessing feature importance rankings, ALE plots characterize the functional form of each predictor’s marginal effect across its empirical distribution. This distinction is particularly critical in ICU settings, where abrupt thresholds or U-shaped response curves—such as those observed for serum creatinine or respiratory parameters—often correspond to actionable clinical inflection points.

Collectively, our approach ensures that the proposed model is not only statistically rigorous and discriminative, but also interpretable in ways that are aligned with clinical reasoning and decision-making. The application of ALE, in particular, represents a methodological strength of this study, enabling the identification of interpretable, locally valid, and clinically relevant feature-risk relationships within a high-dimensional, ICU-specific context.

\section{Results}
\subsection*{Cohort Balance and Clinical Differentiation}

To evaluate the reliability and clinical validity of the mortality prediction model developed for ICU patients with coexisting DM and AF, a stratified statistical comparison was performed. Specifically, we analyzed both the distributional consistency between the training cohort ($n=1074$) and the test cohort ($n=461$), as well as the clinical separability between survivors and non-survivors at 28 days after admission to the ICU.

As summarized in Table~\ref{tab:cohort comparison results}, no significant differences were found in 19 clinical variables when comparing training and test sets, as verified by independent sample $t$-tests with a significance threshold of $p<0.05$.  52.29, $p=0.577$), anion gap (14.64 vs 14.52 mEq / L, $p=0.591$), and BUN (35.40 vs 35.46 mg / dL, $p=0.966$) showed near-identical values, while binary variables such as the proportion receiving extubation (19\% vs. 20\%, $p=0.677$) and those with cardiogenic shock (17\% vs. 12\%, $p=0.006$) also reflected balanced distributions. This indicates that the model training process was not biased by data partitioning and supports the generalizability of the model to unseen ICU admissions.

In contrast, Table~\ref{tab:cohort comparison results 1} highlights pronounced differences in clinical characteristics between the survival and non-survival groups. Patients who died within 28 days exhibited significantly elevated illness severity at the time of admission to the ICU, as reflected in their APS III scores (68.58 vs 48.30, $p<0.001$) and metabolic derangements including higher creatinine (2.63 vs 1.92 mg / dL, $p<0.001$), bilirubin (1.69 vs 0.92 mg / dL, $p=0.000$) and lower albumin (3.02 vs 3.32 g / dL, $p<0.001$). These findings are consistent with multiorgan stress and poor perfusion, factors often exacerbated in AF superimposed on DM.

Importantly, variables that capture physiological deterioration and ICU support burden also demonstrated strong stratification. For example, a higher maximum inspiratory pressure (22.49 vs. 20.02 cmH\textsubscript{2}O, $p<0.001$) and lower Richmond-RAS scale scores ($-2.03$ vs $-0.94$, $p<0.001$) were associated with mortality, suggesting a higher degree of respiratory compromise and sedation needs. The increased incidence of acute respiratory failure with hypoxia (53\% vs. 29\%, $p<0.001$) and severe shock sepsis (50\% vs. 16\%, $p<0.001$) in non-survivors further underscores the layered acute deterioration atop the chronic diabetic background.

These results confirm that, while the training and test sets are methodologically balanced, the selected characteristics provide meaningful clinical stratification. The fusion of DM and acute AF requires nuanced risk modeling, and the observed patterns reinforce the clinical logic behind the model design. This strengthens its potential use in early warning systems for ICU populations with dual chronic-acute comorbidities.

\begin{table}[H]
\noindent
\caption{\textbf{T-test Comparison of Feature Distributions between Training and Test Sets.}}
\label{tab:cohort comparison results}
\small
\renewcommand{\arraystretch}{1.2}
\rowcolors{2}{white}{white}
\begin{tabularx}{\textwidth}{>{\raggedright\arraybackslash}X|c|X|X|X}
\hline
\rowcolor[HTML]{D9EAD3}
\textbf{Feature} & \textbf{Unit} & \textbf{Training Set} & \textbf{Test Set} & \textbf{P-value} \\ \hline
Richmond-RAS Scale & Score & -1.12 (1.35) & -1.14 (1.32) & 0.818 \\ \hline
apsiii & Score & 51.64 (20.38) & 52.29 (21.01) & 0.577 \\ \hline
Anion gap & mEq/L & 14.64 (4.11) & 14.52 (3.93) & 0.591 \\ \hline
INR(PT) & Ratio & 1.61 (0.75) & 1.59 (0.72) & 0.700 \\ \hline
Braden Mobility & Score & 2.45 (0.56) & 2.43 (0.54) & 0.517 \\ \hline
pO2 & mmHg & 128.83 (68.11) & 127.86 (64.98) & 0.792 \\ \hline
Bilirubin, Total & mg/dL & 1.05 (1.48) & 1.13 (2.15) & 0.485 \\ \hline
Creatinine & mg/dL & 2.03 (1.79) & 2.09 (1.76) & 0.563 \\ \hline
Peak Insp. Pressure & cmH\textsubscript{2}O & 20.43 (4.09) & 20.33 (4.00) & 0.681 \\ \hline
Braden Friction/Shear & Score & 2.17 (0.42) & 2.13 (0.43) & 0.091 \\ \hline
BUN & mg/dL & 35.40 (25.14) & 35.46 (23.60) & 0.966 \\ \hline
Albumin & g/dL & 3.27 (0.50) & 3.28 (0.50) & 0.871 \\ \hline
Urea Nitrogen & mg/dL & 35.62 (24.94) & 36.01 (23.77) & 0.772 \\ \hline
age & Years & 69.71 (8.91) & 69.70 (8.90) & 0.986 \\ \hline
Severe sepsis with septic shock & Presence & 0.22 (0.41) & 0.22 (0.41) & 0.967 \\ \hline
Cardiogenic shock & Presence & 0.17 (0.38) & 0.12 (0.32) & 0.006 \\ \hline
Acute respiratory failure with hypoxia & Presence & 0.33 (0.47) & 0.37 (0.48) & 0.150 \\ \hline
insurance & If Medicaid & 0.11 (0.32) & 0.11 (0.31) & 0.908 \\ \hline
Extubation & Presence & 0.19 (0.40) & 0.20 (0.40) & 0.677 \\ \hline
\end{tabularx}
\begin{flushleft}
\textbf{Table notes}: Summary statistics for training and test cohorts across clinical variables. Values are mean (SD). P-values are from t-tests; threshold for significance was set at 0.05.
\end{flushleft}
\end{table}

\begin{table}[H]
\noindent
\caption{\textbf{T-test Comparison of Feature Distributions between Survival and Non-Survival Sets.}}
\label{tab:cohort comparison results 1}
\small
\renewcommand{\arraystretch}{1.2}
\rowcolors{2}{white}{white}
\begin{tabularx}{\textwidth}{>{\raggedright\arraybackslash}X|c|X|X|X}
\hline
\rowcolor[HTML]{D9EAD3}
\textbf{Feature} & \textbf{Unit} & \textbf{Non-Survival} & \textbf{Survival} & \textbf{P-value} \\ \hline
Richmond-RAS Scale & Score & -0.94 (1.16) & -2.03 (1.78) & < 0.001 \\ \hline
apsiii & Score & 48.30 (17.97) & 68.58 (23.28) & < 0.001 \\ \hline
Anion gap & mEq/L & 14.10 (3.56) & 17.40 (5.41) & < 0.001 \\ \hline
INR(PT) & Ratio & 1.55 (0.70) & 1.92 (0.87) & < 0.001 \\ \hline
Braden Mobility & Score & 2.52 (0.53) & 2.10 (0.56) & < 0.001 \\ \hline
pO2 & mmHg & 134.07 (69.71) & 102.25 (51.94) & < 0.001 \\ \hline
Bilirubin, Total & mg/dL & 0.92 (1.16) & 1.69 (2.46) & 0.000 \\ \hline
Creatinine & mg/dL & 1.92 (1.76) & 2.63 (1.80) & < 0.001 \\ \hline
Peak Insp. Pressure & cmH\textsubscript{2}O & 20.02 (3.65) & 22.49 (5.39) & < 0.001 \\ \hline
Braden Friction/Shear & Score & 2.22 (0.40) & 1.91 (0.41) & < 0.001 \\ \hline
BUN & mg/dL & 32.82 (23.65) & 48.47 (28.27) & < 0.001 \\ \hline
Albumin & g/dL & 3.32 (0.47) & 3.02 (0.54) & < 0.001 \\ \hline
Urea Nitrogen & mg/dL & 33.19 (23.65) & 47.96 (27.59) & < 0.001 \\ \hline
age & Years & 69.42 (8.94) & 71.18 (8.62) & 0.014 \\ \hline
Severe sepsis with septic shock & Presence & 0.16 (0.37) & 0.50 (0.50) & < 0.001 \\ \hline
Cardiogenic shock & Presence & 0.14 (0.35) & 0.31 (0.46) & < 0.001 \\ \hline
Acute respiratory failure with hypoxia & Presence & 0.29 (0.46) & 0.53 (0.50) & < 0.001 \\ \hline
insurance & If Medicaid & 0.12 (0.32) & 0.09 (0.29) & 0.270 \\ \hline
Extubation & Presence & 0.22 (0.41) & 0.07 (0.26) & < 0.001 \\ \hline
\end{tabularx}
\begin{flushleft}
\textbf{Table notes}: This table compares patients survive or dead within 28 days. Differences in mean values of key clinical variables are shown along with p-values. Statistical significance set at 0.05 threshold.
\end{flushleft}
\end{table}

\subsection*{Ablation Study and Feature Contribution Analysis}

\begin{figure}[H]
    \centering
    \includegraphics[width=1\linewidth]{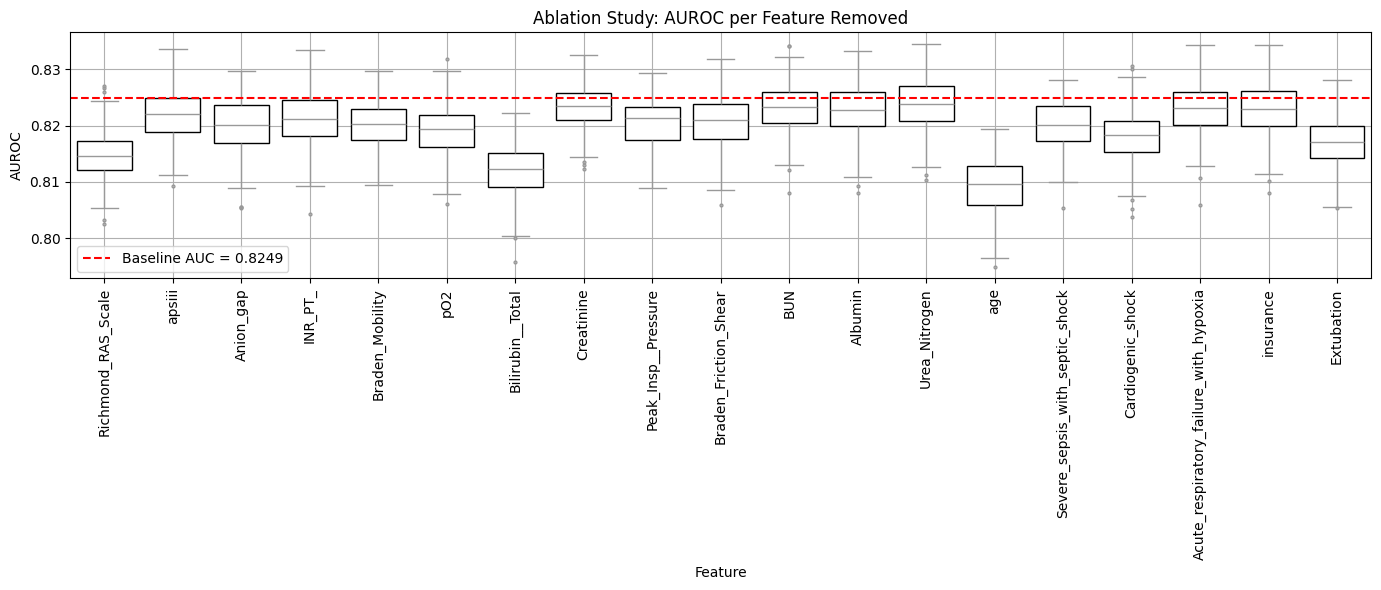}
    \caption{\textbf{Impact of Feature Removal on CatBoost Model Performance.}}
    \label{fig:ablation analysis}
\end{figure}

To examine the robustness and clinical transparency of the logistic regression model developed to predict 28-day mortality in ICU patients with DM and AF, we performed an ablation analysis. As illustrated in Fig.~\ref{fig:ablation analysis}, each clinical characteristic was individually excluded from the input of the model, followed by retraining and bootstrapped evaluation ($n=10$). The AUROC distribution for each experiment is visualized via boxplots, and the red dashed line marks the baseline AUROC of 0.8249 achieved by the full model using all features.

The analysis revealed that the removal of several features—particularly \textit{apsiii}, \textit{age}, \textit{extubation} and \textit{Richmond-RAS Scale}—led to notable reductions in model performance. This underscores their predictive weight and clinical relevance in characterizing the severity, physiological reserve, and neurological responsiveness of the disease in high-risk ICU patients. These features are particularly crucial when evaluating patients with complex chronic-acute interactions, such as AF exacerbating the systemic effects of long-standing diabetes.

Features like \textit{anion gap}, \textit{INR}, and \textit{Braden Friction/Shear} also demonstrated considerable influence, reflecting the role of metabolic acidosis, coagulation abnormalities, and immobility risk in the prognosis of this population. The impact of excluding these variables aligns well with clinical expectations: Patients with impaired perfusion, hemostatic imbalance, or functional deterioration tend to have worse outcomes, and these signs provide early, interpretable warning signals for clinicians.

In contrast, features such as \textit{Urea Nitrogen} showed minimal impact when removed, suggesting that while they may contribute additional nuance, they are not individually critical to the discriminative capacity of the model. However, their presence likely strengthens model performance through multivariate interactions, consistent with the additive nature of logistic regression. Our model is based on tree structure for classification, and the importance of these variables will be reflected in other feature analyses later.

Importantly, the limited variability of the AUROC in most feature removals attests to the stability of the model. The linear nature of the logistic framework also enables intuitive interpretation. Each predictor contributes proportionally to the mortality risk estimate and ablation directly illustrates which predictors have substantive effects.

In general, this feature-level evaluation confirms the biological plausibility and clinical alignment of the core components of the model. By combining interpretable structure with robust performance under systematic perturbation, the model demonstrates strong potential for real-time risk stratification in the ICU setting –particularly among patients facing dual burdens of chronic metabolic disease and acute arrhythmogenic compromise.

\subsection*{Model Performance Evaluation and Clinical Interpretability}

To rigorously evaluate the predictive performance and robustness of machine learning models in estimating 28-day mortality among ICU patients with concurrent DM and AF, we conducted a comprehensive benchmark across seven widely used algorithms. Table~\ref{tab: Results of the Training Set} and Table~\ref{tab: Results of the Test Set} summarize their performance in terms of AUROC, accuracy, sensitivity, specificity, positive predictive value (PPV), negative predictive value (NPV), and F1-score.

\begin{figure}[H]
\centering
\includegraphics[width=1\linewidth]{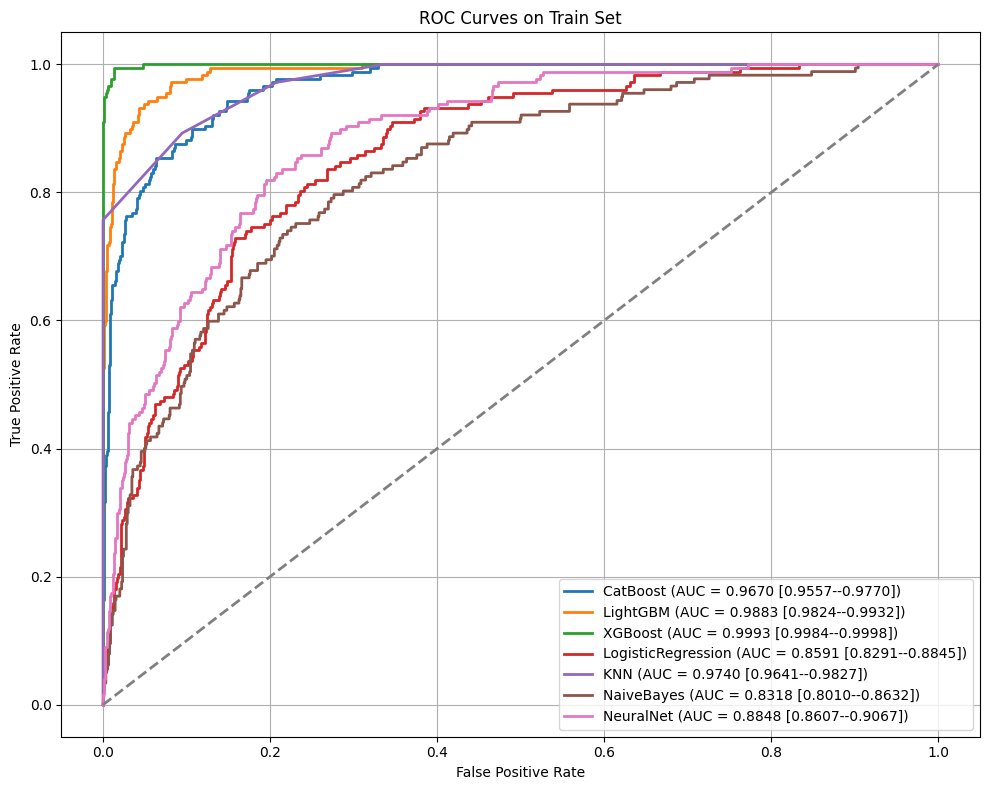}
\caption{\textbf{AUROC Curves for Model Performance in the Training Set.}}
\label{fig:roc_train}
\end{figure}

In the training set(Figure \ref{fig:roc_train}), tree-based ensemble models—particularly XGBoost, LightGBM, and CatBoost—demonstrated strong predictive ability. XGBoost achieved near-perfect performance with an AUROC of 0.999 (95\% CI: 0.998--1.000), closely followed by LightGBM (0.988) and CatBoost (0.967). These models also recorded high accuracy (CatBoost: 0.930; LightGBM: 0.960), reinforcing their capacity to capture intricate nonlinear associations present in heterogeneous ICU data. In contrast, models such as Logistic Regression (AUROC: 0.859) and naive Bayes (AUROC: 0.832) showed relatively lower discriminative ability, reflecting limitations of parametric assumptions in capturing the complex pathophysiological interactions in patients with DM-AF.

\begin{figure}[H]
\centering
\includegraphics[width=1\linewidth]{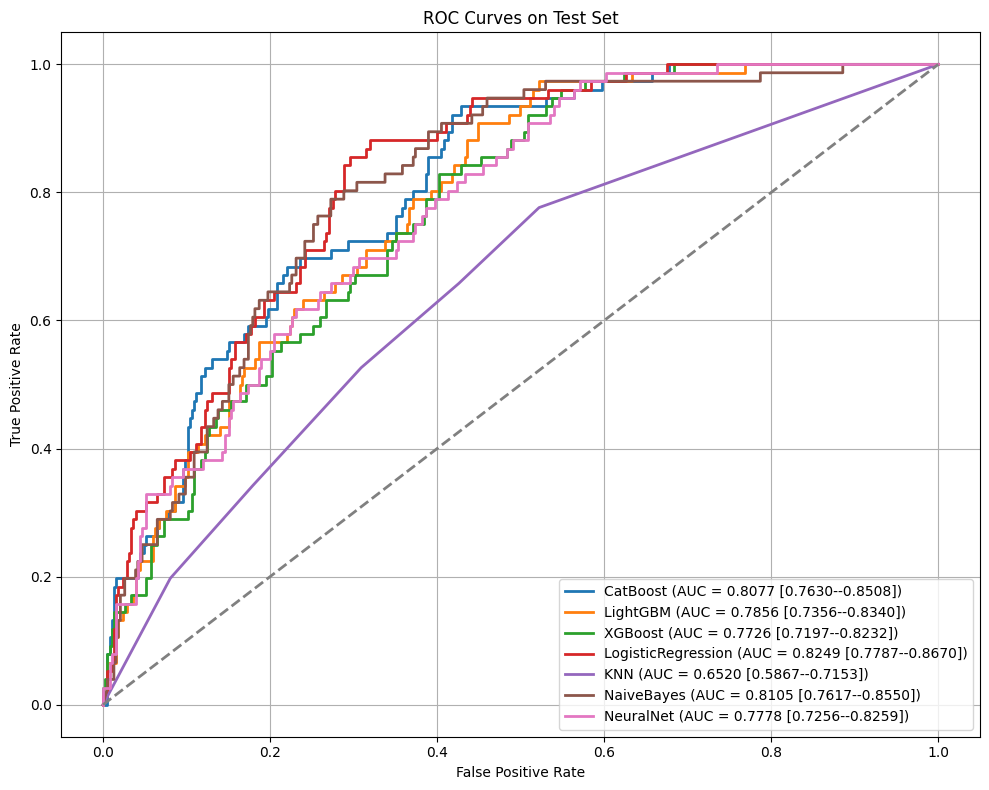}
\caption{\textbf{AUROC Curves for Model Performance in the Test Set.}}
\label{fig:roc_test}
\end{figure}

The performance of the model on the test set(Figure \ref{fig:roc_test}) revealed meaningful insights into generalizability. CatBoost remained competitive with an AUROC of 0.808 (95\% CI: 0.763--0.851), a well-balanced specificity of 0.883, and the highest NPV (0.899) among all models, suggesting clinical utility in ruling out low-risk patients. LightGBM and XGBoost also maintained reasonable AUROCs (0.786 and 0.773, respectively), though they exhibited lower sensitivity and F1 scores, indicating reduced ability to flag true positives. In particular, logistic regression achieved the highest AUROC (0.825) in the test set among all models, with the most balanced sensitivity (0.710) and NPV (0.929), supporting its transparency and resilience when applied to real-world ICU cohorts.

These findings support the value of ensemble learning methods—particularly CatBoost—for predicting clinical risk in settings characterized by multimorbidity and physiological instability. The consistency of CatBoost’s performance, combined with its tolerance for categorical inputs and missing data, positions it as an effective and reliable candidate for ICU deployment. Meanwhile, the surprisingly strong performance of logistic regression outside the sample underscores the advantage of simplicity and interpretability in high-stakes clinical contexts.

From a medical perspective, models that retain performance while allowing interpretability are particularly desirable. Previous ablation analyzes confirmed that these models prioritize clinically coherent features, such as APS III, lactate, and oxygenation metrics, reinforcing the alignment between statistical modeling and medical reasoning.

In general, the results suggest that both CatBoost and Logistic Regression offer viable pathways for the prediction of mortality in ICU patients with DM and AF. CatBoost delivers high accuracy and robust handling of feature interactions, while Logistic Regression offers interpretability and generalization. These properties make them strong candidates for integration into clinical decision support systems that aim to optimize patient triage and improve ICU outcomes.

\begin{table}[H]
\renewcommand{\arraystretch}{1.2}
\centering
\caption{\textbf{Performance Comparison of Different Models in the Training Set.}}
\resizebox{\textwidth}{!}{
\begin{tabular}{l|l|l|l|l|l|l|l}
\hline
\rowcolor[HTML]{D9EAD3}
\textbf{Model} & \textbf{AUROC (95\% CI)} & \textbf{Accuracy} & \textbf{F1-score} & \textbf{Sensitivity} & \textbf{Specificity} & \textbf{PPV} & \textbf{NPV} \\ \hline
CatBoost & 0.967 (0.956--0.978) & 0.930 & 0.789 & 0.798 & 0.956 & 0.783 & 0.960 \\ \hline
LightGBM & 0.988 (0.982--0.993) & 0.960 & 0.875 & 0.855 & 0.981 & 0.900 & 0.971 \\ \hline
XGBoost & 0.999 (0.998--1.000) & 0.989 & 0.966 & 0.960 & 0.994 & 0.971 & 0.992 \\ \hline
\rowcolor[HTML]{FDE9D9}
LogisticRegression & 0.859 (0.829--0.885) & 0.781 & 0.538 & 0.774 & 0.783 & 0.410 & 0.946 \\ \hline
KNN & 0.974 (0.964--0.983) & 0.822 & 0.642 & 0.971 & 0.793 & 0.481 & 0.993 \\ \hline
NaiveBayes & 0.832 (0.801--0.863) & 0.778 & 0.513 & 0.713 & 0.791 & 0.401 & 0.933 \\ \hline
NeuralNet & 0.885 (0.861--0.907) & 0.826 & 0.584 & 0.744 & 0.842 & 0.484 & 0.944 \\ \hline
\end{tabular}
}
\label{tab: Results of the Training Set}
\end{table}

\begin{table}[H]
\small
\renewcommand{\arraystretch}{1.2}
\centering
\caption{\textbf{Performance Comparison of Different Models in the Test Set.}}
\resizebox{\textwidth}{!}{
\begin{tabular}{l|l|l|l|l|l|l|l}
\hline
\rowcolor[HTML]{D9EAD3}
\textbf{Model} & \textbf{AUROC (95\% CI)} & \textbf{Accuracy} & \textbf{F1-score} & \textbf{Sensitivity} & \textbf{Specificity} & \textbf{PPV} & \textbf{NPV} \\ \hline
CatBoost & 0.808 (0.763--0.851) & 0.820 & 0.476 & 0.500 & 0.883 & 0.458 & 0.899 \\ \hline
LightGBM & 0.786 (0.736--0.834) & 0.810 & 0.377 & 0.354 & 0.899 & 0.408 & 0.877 \\ \hline
XGBoost & 0.773 (0.720--0.823) & 0.796 & 0.327 & 0.303 & 0.894 & 0.357 & 0.867 \\ \hline
\rowcolor[HTML]{FDE9D9}
LogisticRegression & 0.825 (0.779--0.867) & 0.746 & 0.479 & 0.710 & 0.752 & 0.362 & 0.929 \\ \hline
KNN & 0.652 (0.587--0.715) & 0.664 & 0.340 & 0.523 & 0.690 & 0.250 & 0.882 \\ \hline
NaiveBayes & 0.811 (0.762--0.855) & 0.750 & 0.482 & 0.713 & 0.758 & 0.368 & 0.931 \\ \hline
NeuralNet & 0.778 (0.726--0.826) & 0.748 & 0.433 & 0.580 & 0.781 & 0.344 & 0.904 \\ \hline
\end{tabular}
}
\label{tab: Results of the Test Set}
\end{table}

\subsection*{ALE-Based Interpretation of Key Predictors}

\begin{figure}[H]
\centering
\includegraphics[width=1\linewidth]{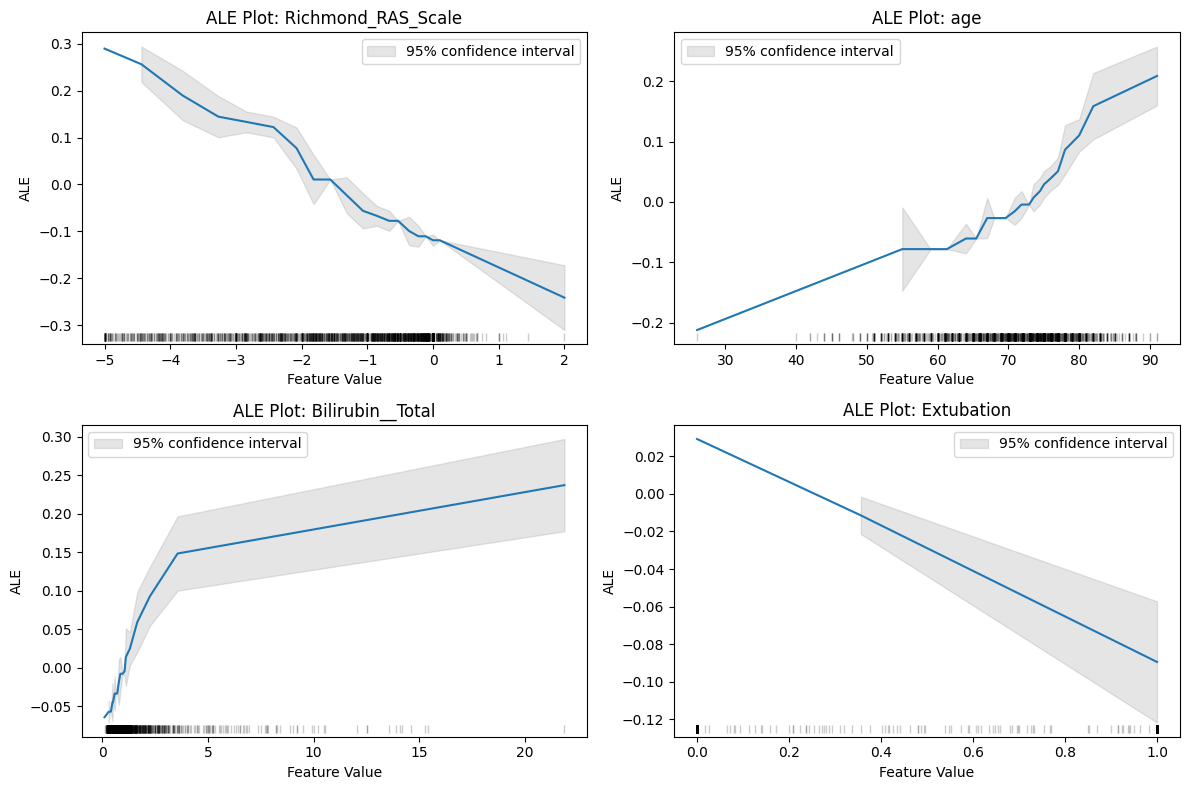}
\caption{\textbf{ALE plots highlight the effects of key clinical features on predicted 28-day mortality in DM-AF patients.}}
\label{fig:ale_analysis}
\end{figure}

To improve the clinical interpretability of the logistic regression model to predict mortality at 28 days among ICU patients with DM and AF, ALE plots were used for four of the most influential characteristics identified in the ablation analysis: Richmond RAS scale, age, total bilirubin, and extubation status (Figure~\ref{fig:ale_analysis}). These features were selected for their consistent and pronounced impact on model performance.

The plot of the Richmond RAS scale demonstrates a clear monotonic downward trend, indicating that lower scores, reflecting deeper levels of sedation or coma, are associated with a significantly increased predicted mortality. This aligns with clinical expectations, where reduced consciousness is often a marker of critical deterioration.

In contrast, the ALE curve for age exhibits an upward slope, particularly beyond 70 years, suggesting a non-linear risk accumulation with advancing age. Although younger patients show relatively stable predictions, the risk increases among the elderly, strengthening the vulnerability of this subgroup.

Total bilirubin reveals a strong positive relationship with mortality risk, especially in the lower range (0–5 mg/dL), beyond which the effect plateaus. This pattern suggests that even moderate liver dysfunction, reflected in early elevation of bilirubin, may substantially influence the prognosis in this population.

Lastly, the extubation variable shows a negative effect, where extubated patients demonstrate a reduced ALE value, reinforcing the notion that successful mechanical ventilation weaning is strongly associated with survival.

Together, these visualizations not only support the model's validity, but also provide intuitive and clinically relevant insights. By focusing on marginal effects independent of multicollinearity, ALE analysis enhances transparency and allows front-line clinicians to better understand which aspects of a patient's physiological and treatment profile are driving risk predictions.

\section{Discussion}
\subsection*{Summary of Existing Model Compilation}
This study introduces a clinically interpretable and statistically robust machine learning framework to predict 28-day mortality in ICU patients with coexisting DM and AF. Our model development pipeline incorporated structured preprocessing, a dual-stage feature selection strategy, and class balancing via SMOTE, followed by comparative evaluation across seven classification algorithms.

Among the models tested, logistic regression achieved the highest generalization performance on the test set (AUROC: 0.825; 95\% CI: 0.779–0.867), outperforming more complex ensemble and neural models. Although often considered simplistic, logistic regression offered strong predictive capacity coupled with improved clinical interpretability, making it highly suitable for high-stakes ICU environments.

Interpretability was achieved through a combination of ablation analysis - which quantified the impact of individual characteristics on model performance - and ALE plots - which revealed linear and clinically significant associations between selected predictors and mortality risk. Dominant predictors such as Richmond RAS scale, age, and total bilirubin emerged as statistically significant and medically relevant, reinforcing the alignment of the model with established clinical reasoning.

By focusing on data from the early phases (first 48 hours after admission to the ICU), our model supports real-time risk stratification for a vulnerable and understudied subpopulation. This has immediate bedside implications, including targeted diagnosis, more aggressive monitoring, and optimized allocation of ICU resources.

\subsection*{Comparison with Prior Studies}

The clinical significance of coexisting DM and AF has been increasingly recognized in the cardiovascular and endocrinology literature. A growing body of evidence demonstrates that the combined presence of these two conditions leads to disproportionately worse outcomes than either condition alone. However, most existing studies have approached this comorbidity from an epidemiological or therapeutic management perspective, and few have developed predictive models—particularly within the context of critical care. Our study directly addresses this gap by offering a machine learning-based, interpretable risk stratification tool focused on short-term ICU mortality in patients with concurrent DM and AF, an area that has remained largely unexplored.

One of the earliest and most cited population-level studies in this domain was conducted by Karayiannides et al.\cite{karayiannides2018high}, who analyzed a nationwide Swedish cohort and found that patients with both AF and DM exhibited significantly higher rates of both cardiovascular and all-cause mortality compared to those with AF alone. The study emphasized that the combination of AF and DM confers an additive risk profile, likely due to synergistic mechanisms such as endothelial dysfunction, prothrombotic states, and autonomic imbalance. While this work offered critical epidemiological insight, it lacked patient-level predictive modeling and was not focused on acute-care settings or ICU admissions, limiting its translational applicability for bedside decision support.

Subsequent research by Liu et al.\cite{liu2024gloria} using data from the GLORIA-AF Phase III registry further reinforced the deleterious effect of diabetes in AF populations. This multinational, prospective registry found that AF patients with diabetes had significantly increased risks of adverse clinical events, including all-cause mortality (HR 1.46), cardiovascular death (HR 1.53), myocardial infarction, and major bleeding, compared to those without diabetes. Interestingly, the risks were even more pronounced among insulin-treated patients, suggesting a gradient of vulnerability based on diabetes severity. However, this study was conducted in ambulatory and stable hospital patients, and although it highlighted the need for personalized care in DM+AF populations, it did not explore acute-phase risk or predictive modeling in the ICU.

A meta-analysis by Xu et al.\cite{xu2022impact} provided further confirmation of the heightened risk associated with this comorbidity. Synthesizing data from over 800,000 patients, the authors demonstrated that AF patients with preexisting diabetes faced significantly higher risks of cardiovascular and all-cause mortality. This analysis added statistical power and generalizability to prior findings but remained focused on long-term outcomes and did not differentiate across levels of clinical acuity, such as ICU care versus general hospitalization or outpatient follow-up. Moreover, it did not propose or validate any predictive algorithms or tools for early risk stratification.

More recently, Geng et al.\cite{geng2022onsetaf} offered compelling evidence of the risk amplification associated with incident AF in individuals with type 2 diabetes. Using UK Biobank data, they followed over 16,000 adults with T2D and demonstrated that those who developed new-onset AF had nearly three times higher all-cause mortality (HR 2.91), fourfold higher cardiovascular death (HR 3.75), and significantly elevated risks of heart failure (HR 4.40), chronic kidney disease (HR 1.68), and atherosclerotic cardiovascular disease (HR 1.85). These findings suggest that AF may act as a catalyst for multi-organ decline in diabetic patients. However, this study was observational, limited to a non-ICU cohort, and did not attempt to construct a clinically deployable prediction model for real-time use.

In addition to these studies, several other reports have described the higher thromboembolic risk, stroke burden, and bleeding complications in DM+AF patients, especially under suboptimal anticoagulation. While these findings collectively underline the prognostic importance of this comorbidity, they offer limited utility for ICU clinicians who require timely, actionable tools to assess patient mortality risk based on early clinical data.

In contrast, our study offers a significant methodological and clinical advancement by developing and validating the first interpretable machine learning model for 28-day ICU mortality prediction specifically in patients with both diabetes and atrial fibrillation. Leveraging early-phase structured data from the MIMIC-IV database, our approach allows for real-time, individualized risk estimation within the first 48 hours of ICU admission. Unlike prior observational work, our pipeline incorporates a two-stage feature selection framework, SMOTE-based class balancing, and robust model interpretability through ablation analysis and ALE. These tools not only enhance predictive performance but also provide clinicians with transparent reasoning behind each risk prediction, thereby improving bedside trust and utility.

By translating population-level findings into an actionable, patient-level prognostic tool for critically ill DM+AF patients, our work fills a critical void in both the machine learning and clinical cardiometabolic literature. It enables timely clinical decisions in a high-risk, underrepresented population—facilitating early intervention, optimized ICU resource allocation, and improved patient outcomes.

\subsection*{Limitations and Future Work}
This study has several limitations. First, despite the use of SMOTE and stratified validation, the moderate sample size of AF+DM patients in MIMIC-IV limits the statistical power and may restrict generalizability to non-U.S. or non-tertiary care settings. Second, while logistic regression was chosen for interpretability, it may fail to capture complex non-linear interactions that could further enhance predictive performance. Future work may explore ensemble or hybrid models that retain interpretability while accommodating higher-order relationships.

Additionally, this model was developed and validated using retrospective data from a single database. External validation across geographically and demographically diverse ICU populations is necessary to ensure clinical generalizability. We also acknowledge the lack of temporal variables and long-term follow-up data, which could enrich predictions of not only 28-day but also 90-day or one-year mortality.

Lastly, while ablation analysis and ALE provide substantial interpretability, integration with clinician-in-the-loop frameworks and prospective implementation in ICU dashboards remains to be tested. Such validation would be critical to assess the true bedside utility of the proposed risk model.

\section{Conclusion}
This study introduces a clinically interpretable and statistically robust machine learning model for predicting 28-day mortality in ICU patients with coexisting DM and AF—a population at uniquely elevated risk due to overlapping metabolic and cardiovascular pathologies. Leveraging data from the MIMIC-IV database, we developed a transparent pipeline integrating a two-stage feature selection strategy, SMOTE-based class balancing, and comparative evaluation across multiple algorithmic frameworks.

Logistic regression outperformed more complex models, achieving a test AUROC of 0.825 (95\% CI: 0.779–0.867). Feature attribution analysis via ablation analysis and ALE consistently identified Richmond-RAS Scale, age, total bilirubin and Extubation as dominant predictors, highlighting the role of illness severity, neurological status, and hepatic function in early prognosis. These insights are clinically intuitive and actionable, reinforcing the model’s utility for bedside decision support.

Compared to prior studies, which either focused on diabetes or AF alone, or relied on long-term observational data, our work is the first to offer an interpretable early-phase mortality prediction tool specifically for ICU patients with this comorbidity. This model not only facilitates individualized risk stratification but also bridges a significant gap in the existing literature.

Future studies should aim to externally validate this framework in non-U.S. cohorts and evaluate its integration into clinical workflows. Expansion to incorporate temporal dynamics and treatment response over time may further enhance predictive performance. Ultimately, this work lays a critical foundation for improving early intervention and optimizing care in one of the ICU’s most vulnerable populations.



\section*{Ethical Statement}

This study was conducted using the publicly available, de-identified MIMIC-IV database (v2.2) developed by the Massachusetts Institute of Technology in collaboration with the Beth Israel Deaconess Medical Center. Use of the MIMIC-IV dataset was approved through the completion of the required data use agreement and ethics training (CITI Program). No further IRB approval was necessary. All data processing and analysis were performed in accordance with relevant guidelines and regulations.

\section*{Author Contributions}

Li Sun led the study design, data analysis, and manuscript drafting and is recognized as the first author. Shuheng Chen, Yong Si, and Junyi Fan contributed equally to data preprocessing, machine learning implementation, and figure preparation. Elham Pishgar provided medical domain expertise and clinical guidance throughout model development and feature interpretation. Kamiar Alaei and Greg Placencia contributed to result interpretation and manuscript editing. Maryam Pishgar supervised the entire project, provided critical revisions, and is the corresponding author.
\section*{Data Availability}

The data used in this study are available through the MIMIC-IV database, which is publicly accessible at https://physionet.org/content/mimiciv/2.2/ upon completion of the data use agreement and required ethics training. No proprietary or patient-identifiable data were used.

\bibliographystyle{elsarticle-num} 
\bibliography{references}







\end{document}